\documentclass{article}

\usepackage{microtype}
\usepackage{graphicx}
\usepackage{subcaption}
\usepackage{booktabs} % for professional tables

\usepackage{hyperref}

\usepackage[preprint]{icml2026}

\usepackage{amsmath}
\usepackage{amssymb}
\usepackage{mathtools}
\usepackage{amsthm}

\usepackage[capitalize,noabbrev]{cleveref}

\theoremstyle{plain}
\newtheorem{theorem}{Theorem}[section]

\newtheorem{lemma}[theorem]{Lemma}

\theoremstyle{definition}

\theoremstyle{remark}

\usepackage[textsize=tiny]{todonotes}

\usepackage{multirow}
\usepackage{pifont}
\newcommand{\cmark}{\ding{51}} % check mark
\newcommand{\xmark}{\ding{55}} % cross mark

\icmltitlerunning{Forward-Only Associative Learning for Test-Time Adaptation}

\begin{document}

\twocolumn[
  % \icmltitle{FAAST: Beyond Backpropagation with Forward-Only Memory-Based Learning}
  \icmltitle{FAAST: Forward-Only Associative Learning via Closed-Form Fast Weights \\ for Test-Time Supervised Adaptation}

  % It is OKAY to include author information, even for blind submissions: the
  % style file will automatically remove it for you unless you've provided
  % the [accepted] option to the icml2026 package.

  % List of affiliations: The first argument should be a (short) identifier you
  % will use later to specify author affiliations Academic affiliations
  % should list Department, University, City, Region, Country Industry
  % affiliations should list Company, City, Region, Country

  % You can specify symbols, otherwise they are numbered in order. Ideally, you
  % should not use this facility. Affiliations will be numbered in order of
  % appearance and this is the preferred way.
  \icmlsetsymbol{equal}{*}

  \begin{icmlauthorlist}
    \icmlauthor{Guangsheng Bao}{zu,wu}
    \icmlauthor{Hongbo Zhang}{zu,wu}
    \icmlauthor{Han Cui}{zu,wu}
    \icmlauthor{Ke Sun}{wu}
    \icmlauthor{Yanbin Zhao}{spu}
    \icmlauthor{Juncai He}{tu}
    \icmlauthor{Yue Zhang}{wu}
  \end{icmlauthorlist}

  \icmlaffiliation{zu}{Zhejiang University}
  \icmlaffiliation{wu}{Westlake University}
  \icmlaffiliation{tu}{Tsinghua University}
  \icmlaffiliation{spu}{Shanghai Polytechnic University}

  \icmlcorrespondingauthor{Yue Zhang}{zhangyue@westlake.edu.cn}
  % \icmlcorrespondingauthor{Firstname2 Lastname2}{zhaoyb553@nenu.edu.cn}

  % You may provide any keywords that you find helpful for describing your
  % paper; these are used to populate the "keywords" metadata in the PDF but
  % will not be shown in the document
  \icmlkeywords{Associative Learning, Fast Weights, Test-Time Supervised Adaptation}

  \vskip 0.3in
]

% this must go after the closing bracket ] following \twocolumn[ ...

% This command actually creates the footnote in the first column listing the
% affiliations and the copyright notice. The command takes one argument, which
% is text to display at the start of the footnote. The \icmlEqualContribution
% command is standard text for equal contribution. Remove it (just {}) if you
% do not need this facility.

% Use ONE of the following lines. DO NOT remove the command.
% If you have no special notice, KEEP empty braces:
\printAffiliationsAndNotice{}  % no special notice (required even if empty)
% Or, if applicable, use the standard equal contribution text:
% \printAffiliationsAndNotice{\icmlEqualContribution}

\begin{abstract}
Adapting pretrained models typically involves a trade-off between the high training costs of backpropagation and the heavy inference overhead of memory-based or in-context learning. We propose FAAST, a forward-only associative adaptation method that analytically compiles labeled examples into fast weights in a single pass. By eliminating memory or context dependence, FAAST achieves constant-time inference and decouples task adaptation from pretrained representation. Across image classification and language modeling benchmarks, FAAST matches or exceeds backprop-based adaptation while reducing adaptation time by over 90\% and is competitive to memory/context-based adaptation while saving memory usage by up to 95\%. These results demonstrate FAAST as a highly efficient, scalable solution for supervised task adaptation, particularly for resource-constrained models. We release the code and models at \url{https://github.com/baoguangsheng/faast}.
\end{abstract}

\section{Introduction}

Backpropagation is the dominant learning paradigm for deep neural networks \cite{rumelhart1986learning} and underpins the success of modern models such as large language models \cite{brown2020language, chowdhery2023palm} and vision-language models \cite{radford2021learning, alayrac2022flamingo}. While highly effective, backpropagation-based adaptation remains expensive in regimes involving many downstream tasks, test-time adaptation, or online learning, where repeated gradient computation, optimizer state maintenance, and iterative updates become a bottleneck \cite{benveniste2012adaptive, finn2017model}. Even parameter-efficient methods such as LoRA \cite{hu2022lora} reduce but do not eliminate these costs, as they still rely on stochastic optimization and GPU-intensive training loops. These limitations motivate alternative adaptation mechanisms that are lightweight, stable, and amenable to rapid deployment.

Recent work has explored memory- or context-based adaptation, which enables models to adapt without parameter updates. In particular, \emph{memory-based methods} store task examples or representations in an external memory and perform explicit lookup at inference time \cite{khandelwal2019generalization, lewis2020retrieval, izacard2023atlas}, while \emph{in-context learning (ICL)} stores task examples in context and allow large language models to perform few-shot learning by conditioning on them \cite{brown2020language}. While effective, these approaches require either external memory or long contexts to hold many examples during inference, which scale at least linearly with the number of the examples \cite{dao2022flashattention, press2021train}. Consequently, existing adaptation strategies either rely on expensive gradient-based optimization or shift the burden to memory/context-dependent inference, motivating mechanisms that are both gradient-free and inference-efficient.

\begin{figure*}[t]
\centering
\includegraphics[trim={0pt 0pt 0pt 0pt},clip,width=1.0\linewidth]{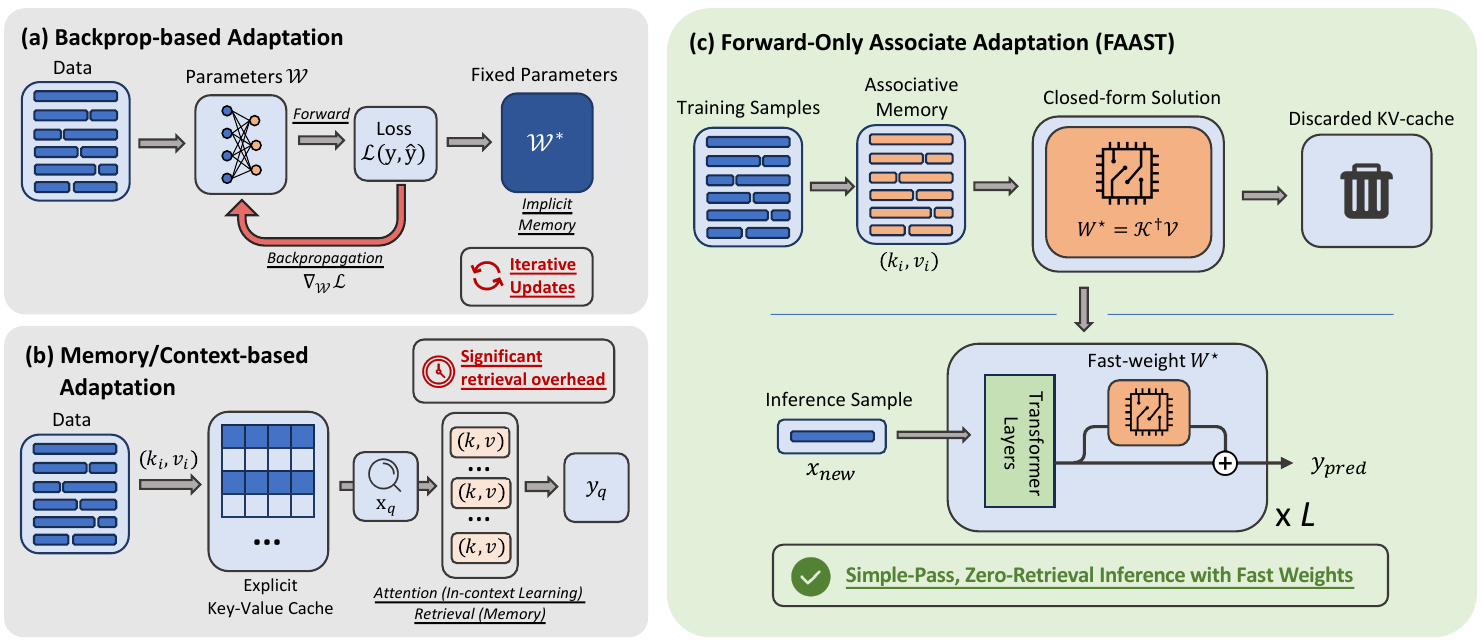}   
\caption{Comparison of downstream task adaptation paradigms.}
\label{fig:intro}
\end{figure*}

In this work, we introduce \textbf{forward-only associative adaptation via spectral transform (FAAST)}, a third regime that avoids both backpropagation and context-length-dependent inference costs. The central observation is that downstream task adaptation often does not require modifying the representation of concepts or objects, but rather learning an \textbf{associative mapping} between pretrained input and output embeddings \citep{he2021effectiveness,wang2025parameter,bourigault2025frevl}. This motivates us to decompose learning into two parts:
(1) \emph{representation learning}, handled by pretrained encoders and kept fixed during adaptation \cite{howard2018universal,devlin2019bert}, and
(2) \emph{associative learning}, which maps input representations to output representations in a task-specific manner, reminiscent of classical associative memory and fast-weight models \cite{hebb1949organization, hinton1987fast, ba2016using}.
FAAST realizes forward-only associative learning by compiling associative memory (paired inputs and outputs) in the form of key-value pairs into \textbf{fast weights} by solving a linear regression problem in closed form.

Figure~\ref{fig:intro} compares the three adaptation paradigms discussed above.
Figure (a) shows \emph{backpropagation-based adaptation}, where task-specific associations are encoded as learned weights via iterative gradient descent. 
Figure (b) illustrates \emph{memory- or context-based adaptation}, which injects task information through memory lookup or in-context attention at inference time, incurring costs that scale with the number of examples. 
Figure (c) presents \emph{FAAST}, which compiles labeled key-value pairs from frozen encoders into fast weights, enabling single-pass, gradient-free learning and constant-cost inference.

FAAST is a non-parametric module, which can be embedded into existing neural networks that produce meaningful representations. Typically, in the context of large language models, we use successive hidden states of all tokens from middle layers as keys and values. New associations between context and desired outputs are appended into the memory and compiled into a projection matrix. At inference time, the model conditions on both its original parametric knowledge and the learned fast weights. Unlike ICL, which requires retaining the full demonstration context in the attention cache, FAAST only preserves the computed fast weights, allowing the stored k–v pairs to be discarded after learning and resulting in substantially lower memory usage.

We evaluate FAAST on typical supervised learning tasks including classification tasks and sequence modeling tasks. On image classification benchmarks, we show that FAAST achieves the same level of accuracy as backprop-based adaptation while saving \textbf{95\%} learning time. On language modeling tasks, FAAST enables small language models such as GPT-2 to have test-time adaptation ability, while saving more than \textbf{93\%} training and inference cost than memory/context-based adaptation. On natural language downstream tasks, including sentiment classification tasks and sequence-to-sequence machine translation tasks, FAAST achieves consistently  better full-set performance compared to LLM zero-shot or ICL few-shot baselines.

\begin{table*}[t]
\centering\scriptsize
\caption{Comparison of FAAST with Representative Prior Approaches.
Symbols: {\cmark= Yes}, {\xmark = No}, {$\triangle$ = Partial}, {$\diamond$ = Rare}.}
\label{tab:related_work}
\resizebox{\textwidth}{!}{
\begin{tabular}{l c p{8.5cm} c}
\hline
\textbf{Property} & \textbf{Prior Work} & \textbf{Prior Representative Examples} & \textbf{FAAST} \\
\hline
Strict representation/association separation
& $\triangle$
& Linear probes \citep{alain2016understanding}; World models \citep{ha2018world}; Actor--critic RL \citep{sutton1998reinforcement} --- separation assumed but gradients or task losses often still influence representations
& \cmark\\

Forward-only associative learning
& \cmark
& Perceptron / LMS \citep{rosenblatt1958perceptron,widrow1988adaptive}; Hebbian learning \citep{hebb1949organization}; TD learning with fixed features \citep{sutton1988learning}
& \cmark\\

Closed-form associative learning
& $\diamond$
& Linear regression heads \citep{kornblith2019better}; Extreme Learning Machines \citep{huang2006extreme}; Ridge regression probes \citep{alain2016understanding}
& \cmark\\

Non-parametric memory
& \cmark
& kNN / episodic memory \citep{blundell2016model}; kNN-LM \citep{khandelwal2020generalization}; RAG \citep{lewis2020retrieval}
& \cmark\\

No prediction error signal
& \xmark
& Supervised learning \citep{rumelhart1986learning}; Hebbian rules \citep{hebb1949organization}; Contrastive learning \citep{oord2018representation}; RL / TD learning \citep{sutton1988learning} --- error signals is generally required
& \cmark\\

No inference-time memory access
& \xmark
& kNN \citep{cover1967nearest}; episodic control \citep{blundell2016model}; RAG \citep{lewis2020retrieval}; Memory Networks \citep{weston2014memory} --- memory queried at inference
& \cmark\\

Fast weights from memory
& $\triangle$
& Fast weights \citep{ba2016using}; Hebbian fast weights \citep{miconi2018differentiable} --- learned via online updates, not analytic derivation
& \cmark\\

Plug-in for pretrained models
& $\triangle$
& Adapters \citep{houlsby2019parameter}; LoRA \citep{hu2022lora}; Retrieval augmentation \citep{borgeaud2022improving}
& \cmark\\

LLM adapt w/o finetuning or retrieval
& \xmark
& RAG \citep{lewis2020retrieval}; kNN-LM \citep{khandelwal2020generalization}; Prompt tuning \citep{lester2021power} --- retrieval or finetuning required
& \cmark\\
\hline
\end{tabular}
}
\end{table*}

In summary, we contribute:
\begin{itemize}
    \item We propose \emph{forward-only associative adaptation},  formalizing task adaptation as a forward-only associative learning process that avoids backpropagation, gradient descent, iterative updates, and prediction-error signals.
    \item We introduce \emph{closed-form fast-weight construction}, compiling key-value pairs into task-specific fast weights in closed form, allowing the memory to be discarded at inference time.
    \item We demonstrate that FAAST enables \emph{plug-and-play task adaptation for pretrained models}, where the module can be embedded as a modular component in pretrained networks, including large language models.
\end{itemize}

\section{Related Work}

As Table~\ref{tab:related_work} summarizes, FAAST is related to a broad line of work on associative learning, fast weights, and alternatives to gradient-based adaptation. Prior studies have explored individual components such as associative memory, frozen representations, biologically inspired forward-only learning rules, and pseudoinverse-based solutions. For example, linear probes and world models separate representation learning from task-specific prediction \citep{alain2016understanding,ha2018world}, while fast-weight and Hebbian-style models provide mechanisms for rapid association \citep{hebb1949organization,schmidhuber1992learning,ba2016using}. However, these approaches typically rely on iterative updates, learned plasticity rules, or continued gradient-based optimization. FAAST differs by enforcing a strict architectural separation in which associative learning operates analytically on fixed pretrained representations, enabling single-pass, optimizer-free adaptation.

FAAST is also closely related to work on forward-only and biologically motivated learning rules that seek to avoid backpropagation, including feedback alignment and forward-forward methods \citep{lillicrap2016random,hinton2022forward}. While these methods demonstrate that learning without error backpropagation is possible, they are generally designed for training representations from scratch or require multiple forward passes and specialized objectives. In contrast, FAAST targets downstream task adaptation on pretrained models, computing task-specific fast weights in closed form with deterministic guarantees. Compared to recent forward-only or zeroth-order adaptation methods \citep{malladi2023fine}, FAAST avoids stochastic search and instead leverages analytic associative memory to achieve efficient and stable adaptation.

Finally, FAAST differs fundamentally from parameter-efficient fine-tuning and memory-based adaptation methods. Techniques such as adapters, LoRA, and prefix tuning reduce training cost but still depend on gradient-based optimization \citep{houlsby2019parameter,hu2022lora,li2021prefix}. In-context learning and memory-augmented models adapt behavior at inference time by conditioning on or querying stored examples \citep{brown2020language,khandelwal2020generalization,lewis2020retrieval}, incurring memory access and attention overhead during inference. FAAST instead compresses all task-specific associations into a single fast-weight matrix, eliminating inference-time memory access while retaining non-parametric storage and rapid adaptation. A detailed comparison with these lines of work is provided in Appendix~\ref{app:related_work}.

\begin{figure*}[t]
    \centering
    \includegraphics[trim={0pt 0pt 0pt 0pt},clip,width=1.0\linewidth]{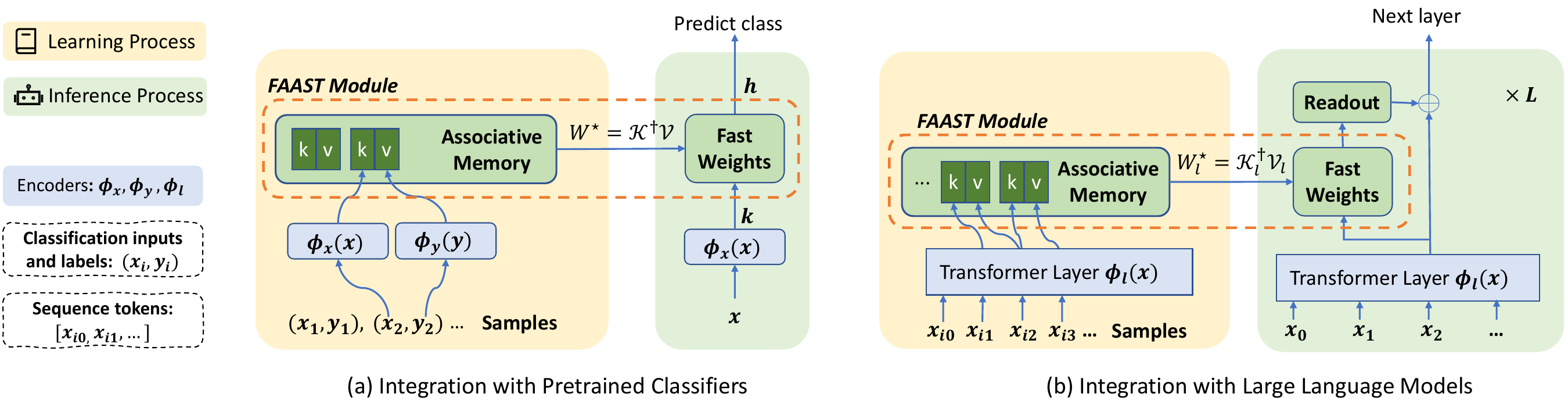}   
    \caption{FAAST module and the integration with pretrained neural networks.}
    \label{fig:method}
\end{figure*}

\section{Preliminaries}
\label{sec:preliminaries}

\subsection{Problem Setup and Notation}

We consider supervised adaptation tasks defined by a dataset
\begin{equation}
\mathcal{D} = {(x_i, y_i)}_{i=1}^N,
\end{equation}
where $x_i \in \mathcal{X}$ denotes an input instance (e.g., image or text), and $y_i \in \mathcal{Y}$ denotes a supervision signal (e.g., class label in classification or next token in sequence modeling).

We assume access to pretrained and frozen encoders \citep{devlin2019bert,radford2021learning}
\begin{equation}
\phi_x: \mathcal{X} \rightarrow \mathbb{R}^{d_x}, \qquad
\phi_y: \mathcal{Y} \rightarrow \mathbb{R}^{d_y},
\end{equation}
which map inputs and outputs into fixed-dimensional embedding spaces.
Thus, each labeled example induces a \textbf{key-value pair}:
\begin{equation}
\mathbf{k}_i = \phi_x(x_i), \qquad \mathbf{v}_i = \phi_y(y_i).
\end{equation}
The task is to learn associations between the keys and values represented in embedding spaces.

\subsection{Task Adaptation via Backpropagation}

A simple downstream adaptation learns a linear projection $W$  that maps input embeddings $\mathbf{k}_i$ to output embedding space \citep{alain2016understanding,kornblith2019better}:
\begin{equation}
\mathbf{h}_i = \mathbf{k}_i^\top W, \qquad
W \in \mathbb{R}^{d_x \times d_y}
\end{equation}
where we assume embeddings are normalized and therefore omit the bias term.

For classification problems, the $y$ is a class label, which probability is computed via an attention head
\begin{equation}
p(y \mid x_i) =
\frac{\exp(\mathbf{h}_i^\top \mathbf{v}_y)}
{\sum_{c=1}^K \exp(\mathbf{h}_i^\top \mathbf{v}_c)}.
\end{equation}

The projection matrix $W$ is learned by minimizing cross-entropy loss using gradient-based optimization.

This linear projection functions as an \textbf{implicit associative memory} \citep{hopfield1982neural,hinton1987fast,ba2016using}, encoding task-specific associations in its parameters. However, learning requires iterative backpropagation and must be repeated for each downstream task.

For sequence modeling tasks such as language modeling \citep{bengio2003neural,vaswani2017attention}, the input $x_i$ corresponds to a contextual token sequence, and the supervision $y_i$ is the next token to be predicted. The same linear projection and softmax formulation applies, with output embeddings representing the vocabulary tokens.

\subsection{Task Adaptation via Memory or ICL}

Task adaptation can also be achieved by \textbf{storing and retrieving labeled examples}. Each training instance is represented as a key-value pair in an explicit memory or input context, and predictions are produced by retrieving relevant values for a query input using similarity-based matching \citep{cover1967nearest} in memory-based methods or implicitly retrieved through self-attention \citep{brown2020language} in ICL.
Generally, given a query representation $\mathbf{q}$, attention-based memory or context models retrieve an output via
\begin{equation}
\label{eq:attention}
\mathbf{h} = \mathbf{a}^\top V, \qquad \mathbf{a} = \mathrm{Attn}(\mathbf{q}, K),
\end{equation}
where $K$ and $V$ are matrices of the keys and values. In this attention-based retrieval, larger attention weights correspond to memory items that are more relevant to the query.

\section{Method}
\label{sec:method}

We propose a forward-only associative learning architecture that enforces a strict separation between \emph{representation learning} and \emph{associative learning}. 
Unlike prior fast-weight approaches \citep{ba2016using,hinton1987fast}, FAAST computes fast weights analytically in closed form, yielding a deterministic, single-pass solution.
We illustrate the basic FAAST module in Section \ref{sec:faast_module} and describe how it can be integrated into existing neural networks in Section \ref{sec:faast_integration}.

\subsection{FAAST Module}
\label{sec:faast_module}

The core of many downstream tasks, including classification and sequence prediction, lies in an associative mapping from an input representation to a corresponding output representation. The key insight is that, once representations are fixed, the optimal linear function representing the associative mapping can be \textbf{computed directly from the stored key-value pairs}, instead of a numerical approximation discovered by stochastic gradient descent.

Formally, given a dataset $\mathcal{D} = \{(x_i, y_i)\}_{i=1}^N$, we collect the key-value pairs into matrices
\begin{align}
K &= [\mathbf{k}_1, \dots, \mathbf{k}_N]^\top \in \mathbb{R}^{N \times d_x}, \\
V &= [\mathbf{v}_1, \dots, \mathbf{v}_N]^\top \in \mathbb{R}^{N \times d_y}.
\end{align}
FAAST defines the task-specific associative mapping as a \textbf{fast-weight matrix} computed by solving the linear regression problem
\begin{equation}
\min_W \| K W - V \|_F^2,
\end{equation}
The optimal solution is given analytically by
\begin{equation}
W^\star = K^\dagger V \in \mathbb{R}^{d_x \times d_y}
\end{equation}
where $K^\dagger$ denotes the Moore-Penrose pseudoinverse \citep{penrose1955generalized}.

The Moore-Penrose pseudoinverse can be solved by singular value decomposition (SVD), a spectral transform of matrix. Specifically, let the SVD of $K$ be
\begin{equation}
K = \mathcal{U}\, \Sigma\, \mathcal{R}^\top,
\end{equation}
where singular values $\Sigma = \mathrm{diag}(\sigma_1, \dots, \sigma_r)$ with $\sigma_1 \ge \dots \ge \sigma_r > 0$, and singular vectors $\mathcal{U}$ and $\mathcal{R}$ have orthonormal columns.
The pseudoinverse is then given by
\begin{equation}
K^\dagger = \mathcal{R}\, \Sigma^\dagger\, \mathcal{U}^\top, \qquad
\Sigma^\dagger = \mathrm{diag}(\sigma_1^{-1}, \dots, \sigma_r^{-1}),
\end{equation}
and the fast weights can be written as
\begin{equation}
W^\star = \mathcal{R}\, \Sigma^\dagger\, \mathcal{U}^\top V.
\end{equation}

The computation of $W^\star$ involves only a \textbf{single forward pass over the data} and yields a deterministic solution with theoretical optimality (see Appendix \ref{app:fastweights_regression}).

\paragraph{Incremental Update Rule.}
A key challenge in supervised adaptation is scale: while classification tasks may involve up to $10^6$ input-output pairs, language models may involve the order of $10^{10}$ tokens. Storing all key-value pairs explicitly is infeasible. To address it, we propose the incremental update rule of the fast-weight matrix:
\begin{equation}
\label{eq:incremental_update}
\small
W_{t+1}
=
\frac{N_t}{N_{t+1}} W_t
+
\frac{N}{N_{t+1}} W^\star,
\qquad
N_{t+1} = N_t + N,
\end{equation}
where $W^\star$ is computed from a new batch of $N$ key-value pairs and update the new weights $W_{t+1}$. This update rule incrementally aggregates associative evidence without retaining all past data, and its validity is theoretically justified in Appendix~\ref{app:incremental_update}.

\paragraph{Trade-off Between Underfitting and Overfitting.}

The generalization behavior of the fast-weight matrix $W^\star$ is governed by the spectral structure of the key matrix $K$. 
Each singular component of $K$ contributes independently to the associative mapping. \emph{Large singular values} capture dominant directions in the data and tend to encode task-relevant structure that generalizes across samples. In contrast, \emph{small singular values} correspond to poorly supported directions; amplifying these directions via $\sigma_i^{-1}$ can lead to memorization of noise or idiosyncratic examples. 

This observation motivates the use of spectral filtering, providing explicit, interpretable trade-offs between underfitting and overfitting. By truncating singular values below a relative threshold $\epsilon$, we suppress unstable components. Such filtering prevents overfitting in low-data regimes while retaining task-relevant directions in larger datasets, ensuring stable generalization. Unlike Ridge Regression \citep{hoerl1970ridge}, which shrinks all directions uniformly, spectral filtering directly targets task-relevant components. Formally, we define a filtered pseudoinverse \citep{van1996matrix}:
\begin{equation}
\Sigma^\dagger_\epsilon = \mathrm{diag}\big(\sigma_i^{-1}\,\mathbb{I}[\sigma_i \ge \sigma_{\text{max}} \epsilon]\big),
\end{equation}
which is computed entirely from forward memory statistics. In practice, we set $\epsilon=1/N^\alpha$, where $N$ is the number of key-value pairs and $\alpha\in[0,1]$ reflects task complexity, with a default of 1.

\paragraph{Fast Weights as Pseudoinverse Attention.}
The closed-form fast-weight solution $W^\star$ admits an interpretation as an attention-based retrieval mechanism that solves a least-squares matching problem between queries and stored keys. This \emph{pseudoinverse attention} computes signed attention weights $\mathbf{a}^\star = K^\dagger \mathbf{q}$, yielding the retrieved output $\mathbf{h} = (\mathbf{a}^\star)^\top V$, and thus enables both additive and subtractive interactions beyond convex combinations. From this perspective, FAAST represents a fully compressed limit of attention-based memory with no inference-time memory access. We further show that softmax attention arises as an entropy-regularized relaxation of pseudoinverse attention; see Appendix~\ref{app:pseudoinverse_attn} for details.

% \paragraph{Inference Without Memory Access.}
% At inference time, given input embedding $\mathbf{k}=\phi_x(x)$, prediction reduces to a single matrix multiplication:
% \begin{equation}
% \mathbf{h} = \mathbf{k}^\top W^\star.
% \end{equation}
% Crucially, only the fast weights $W^\star$ are required. The original key-value pairs can be discarded once $W^\star$ is computed. This sharply distinguishes FAAST from memory-based methods, which require storing and querying all examples at inference time.

% For a new task, FAAST simply recomputes $W^\star$ from the new task’s key-value pairs. Because pretrained parameters are never modified, the same backbone can be rapidly adapted to many tasks, making the method well suited for few-shot, many-task, and test-time adaptation regimes.

\begin{table*}[t]
\centering\small
\caption{Image classification results about 5-shot and full-data accuracy with 95\% confidence interval for CIFAR10  and {\it mini}ImageNet. Inference FLOPs and memory usage only count projection layer, assuming $d_x=d_y=1024$ and $N=10{,}000$.}
\label{tab:image_classification_main}
\begin{tabular}{lc|cc|cc|ccc}
\toprule
\multirow{2}{*}{\bf Method} 
& \multirow{2}{*}{\bf Adaptation} 
& \multicolumn{2}{c}{\bf CIFAR10 (10-way)} 
& \multicolumn{2}{c}{\bf {\it mini}ImageNet (20-way)}  
& \multicolumn{2}{c}{\bf + Inference Cost} & \bf Learn Cost \\
 
&  
& 5-shot 
& Full 
& 5-shot 
& Full 
& Compute 
& Memory & GPU Secs \\

\midrule
CLIP (lower bound) 
& No adapt
& \multicolumn{2}{c|}{70.3$\pm$0.9} 
& \multicolumn{2}{c|}{86.7$\pm$0.9} 
& $\mathcal{O}(d_x)$ & 0 MB
& - \\

Linear (upper bound)
& Backprop
& \color{red} 71.9$\pm$0.4 & \bf 88.3$\pm$0.3 
& \color{red} 84.5$\pm$0.3 & \bf  93.2$\pm$0.2
& $\mathcal{O}(d_x^2)$ & 4 MB
& 780 s \\

Full Finetuning
& Backprop
& \color{red} 13.3$\pm$3.4  & \color{red} 85.4$\pm$0.4 
& \color{red} 6.9$\pm$3.1 & \color{red} 87.5$\pm$0.6  
& $\mathcal{O}(d_x^2)$ & 4 MB
& 1,603 s \\

k-NN Memory 
& Memory
& 71.7$\pm$0.3 & 82.7$\pm$0.2 
& 87.1$\pm$0.2 & 88.6$\pm$0.1 
& $\mathcal{O}(N d_x)$ & 80 MB
& \multirow{2}{*}{\bf 41 s}  \\

Softmax Memory
& Memory
& 71.5$\pm$0.3 & 80.5$\pm$0.2 
& 87.7$\pm$0.2 & 87.9$\pm$0.2 
& $\mathcal{O}(N d_x)$ & 80 MB 
& \multirow{2}{*}{\bf (-94.7\%)} \\

\bf FAAST (Ours)
& Fast weights
& \bf 73.8$\pm$0.3 & \bf 86.7$\pm$0.2 
& \bf 88.6$\pm$0.2 & \bf 93.0$\pm$0.1 
& \bf $\mathcal{O}(d_x^2)$ & \bf 4 MB 
&  \\
\bottomrule
\end{tabular}
\end{table*}

\subsection{FAAST Integration into Pretrained Networks}
\label{sec:faast_integration}

FAAST is designed as a plug-in associative learning module that can be integrated into existing neural networks to enable efficient downstream adaptation. Below, we illustrate this integration for two representative models: pretrained neural classifiers and language models.

\paragraph{Pretrained Neural Classifiers.}

Integrating FAAST into a pretrained classifier is straightforward. Consider a classifier with a frozen backbone that produces representations $\phi_x(x)$ and $\phi_y(y)$, and an original output layer parameterized by a projection matrix $W_0$.
Instead of replacing this pretrained projection, we linearly interpolate it with the FAAST projection $W^\star$, following Eq.~\ref{eq:incremental_update}. Here, $N_0$ denotes the effective sample size associated with the pretrained projection, and $N$ is the number of key-value pairs used to construct $W^\star$. As the memory size $N$ increases, the resulting classifier smoothly transitions from prior-dominated predictions to task-specific adaptation.

% This interpolation admits a natural Bayesian interpretation as a shrinkage estimator~\citep{hastie2009elements}. The pretrained projection $W_0$ serves as a prior, while FAAST contributes data-driven evidence from the associative memory. As the memory size $N$ increases, the resulting classifier smoothly transitions from prior-dominated predictions to task-specific adaptation.

\paragraph{Pretrained Language Models.}

The integration of FAAST into sequence models and large language models follows the same general principle, with additional considerations arising from scale and temporal structure. Formally, given an input sequence $x = (x_1, \dots, x_T)$, we extract hidden representations from intermediate layers of a pretrained transformer \citep{vaswani2017attention}:
\[
\mathbf{k}_{\ell,t} = \phi_\ell(x)_t,
\qquad
\mathbf{v}_{\ell,t} = \phi_\ell(x)_{t+1},
\]
which form key-value pairs associating past token representations with future ones. These pairs are aggregated across time steps forming $K_l$ and $V_l$, and compressed into a fast-weight matrix $W^\star_l$ per layer.

To interface the memory output with pretrained Transformer layers, we employ a residual connection together with a lightweight linear readout projection $P_\ell$:
\begin{equation}
\mathbf{h}_{\ell,t} = \mathbf{k}_{\ell,t} + \mathbf{k}_{\ell,t}^\top W_l^\star P_\ell,
\end{equation}
where $P_\ell$ is initialized with zero weights to avoid intrusion to existing fitting between layers and is trained on diverse texts.

The readout projection $P_\ell$ is task-independent, and the product $W_\ell^\star P_\ell$ can be folded into a single matrix at inference time. The readout is trained once to map memory-adapted representations back into the input space of the subsequent Transformer layer and is kept fixed during downstream adaptation. All task-specific learning is thereafter captured solely by the fast weights $W_\ell^\star$.

Finally, since not all tokens contribute equally to future prediction, we incorporate a lightweight key--value importance scorer. The scorer is implemented as a linear classifier over the concatenation of $\mathbf{k}$ and $\mathbf{v}$, followed by a sigmoid activation to produce weights in $[0,1]$. Trained jointly with the readout projection, these weights modulate the contribution of individual key-value pairs during fast-weight construction, enabling the memory to emphasize informative associations while suppressing noise.

\section{Experiments on Supervised Classification Tasks}

We test whether effective adaptation can be achieved by learning associative mappings over fixed representations. The supervised classification benchmarks enable a direct comparison between FAAST, gradient-based adaptation, and memory/context-based adaptation across multiple modalities.

\subsection{Image Classification}
\label{sec:imgcls}

Image classification provides a clean testbed for downstream adaptation, as high-quality representations can be obtained from pretrained encoders.

\paragraph{Settings.}
Our image classification experiments utilize a frozen \emph{CLIP ResNet-50} backbone \citep{radford2021learning}, where fixed image and text embeddings serve as keys and values. We evaluate our approach against several baselines, including \emph{CLIP zero-shot} \citep{goh2021multimodal}, \emph{linear projection} \citep{kolesnikov2019revisiting}, \emph{full finetuning}, \emph{k-NN memory} \citep{wu2018unsupervised}, and \emph{softmax memory} \citep{vaswani2017attention}, all operating on identical features to isolate the effects of the memory mechanism. Testing is performed on \emph{CIFAR-10} \citep{krizhevsky2009learning} and \emph{mini-ImageNet} \citep{vinyals2016matching} datasets across both few-shot episodic and full-data regimes. All hyperparameters and backpropagation training configurations are standardized to ensure a fair comparison. For a comprehensive breakdown of the baseline implementations and specific training hyperparameters, please refer to Appendix \ref{app:image_settings}.

\begin{table}[t]
\centering\small
\caption{Sentiment classification accuracy on SST-2 and IMDB datasets, with a 95\% confidence interval.}
\label{tab:gpt2_sentiment_results}
\begin{tabular}{lccc}
\toprule
\textbf{SST2 $\rightarrow$} 
& \textbf{$1$-shot} 
& \textbf{$5$-shot} 
& \textbf{full} \\

GPT2-XL (zero-shot)
& - & - & 74.3$\pm$1.2 \\

In-Context Learning
& 59.6$\pm$1.4 
& 71.6$\pm$1.3
& - \\

\textbf{FAAST (Ours)} 
& \bf 78.5$\pm$1.1
& \bf 80.8$\pm$1.1
& \bf 87.5$\pm$0.9 \\

\midrule
\textbf{IMDB $\rightarrow$} 
& \textbf{$1$-shot} 
& \textbf{$2$-shot} 
& \textbf{full} \\

GPT2-XL (zero-shot)
& - & - & 85.7$\pm$1.0 \\

In-Context Learning
& 70.1$\pm$1.3 
& 56.0$\pm$1.4
& - \\

\textbf{FAAST (Ours)} 
& \bf 86.7$\pm$0.9
& \bf 87.4$\pm$0.9
& \bf 90.4$\pm$0.8 \\
\bottomrule
\end{tabular}
\end{table}

\begin{table*}[t]
\centering\small
\caption{Sequence modeling adaptation results on WikiText-103 using GPT2-XL (1.5B). Inference FLOPs and memory only count increased cost upon GPT2-XL base model, using hidden size $d_x=1600$, number of layers $L=48$, and number of train set tokens $N=1.03\times 10^8$. $^\heartsuit$A theoretical estimation given the base model does not support such long context.}
\label{tab:gpt2_ppl_results}
\begin{tabular}{lcccccc}
\toprule
\multirow{2}{*}{\bf Method} 
& \multirow{2}{*}{\bf Adaptation} 
& \textbf{Test-time} 
& \textbf{WikiText-103}
& \multicolumn{2}{c}{\bf + Inference Cost} & \bf Learn Cost \\

&
& \textbf{Learning}
& PPL $\downarrow$
& Compute
& Memory & GPU hrs \\

\midrule
GPT2-XL (zero-shot, lower bound)
& No adapt
& \xmark
& 17.41
& -
& - 
& - \\

Linear Projection (upper bound)
& Backprop
& \xmark
& 13.60
& $\mathcal{O}(L d_x^2)$
& \bf 112 MB 
& 3 hrs\\

LoRa w/ same number of params
& Backprop
& \xmark
& 13.57
& -
& - 
& 3 hrs \\

In-Context Learning
& Context
& \cmark
& not applicable
& $\mathcal{O}(N L d_x) ^ \heartsuit$ 
& 29 TB $^\heartsuit$ 
& - \\

$k$NN-LM
& Memory
& \cmark
& \bf 12.70
& $\mathcal{O}(N d_x)$
& 307 GB
& 16 hrs \\

\textbf{FAAST (Ours)} 
& Fast weights 
& \cmark
& 15.35
& $\mathcal{O}(L d_x^2)$
& \bf 112 MB 
& \bf 0.2 hrs \\

\bf FAAST w/ readout seen WikiText103
& Fast weights 
& \cmark
& 13.23
& $\mathcal{O}(L d_x^2)$
& \bf (-99.9\%) 
& \bf (-93.3\%) \\

\bottomrule
\end{tabular}
\end{table*}

\paragraph{Results.}
We compare FAAST with backprop-trained linear projection, contrasting gradient-based optimization with closed-form fast weights. We report classification accuracy together with inference computation and memory usage, isolating the cost of associative learning by accounting only for the projection layer (Table~\ref{tab:image_classification_main}). This setup evaluates whether competitive adaptation is possible without gradients, optimizer state, or multiple training epochs.

FAAST consistently outperforms CLIP zero-shot baselines and improves smoothly from few-shot to full-data regimes. Compared with backprop-based adaptations, FAAST is more robust in low-data settings, where linear probing and full finetuning tend to overfit, while remaining competitive at scale. Moreover, FAAST generalizes beyond pretrained semantic priors, achieving high accuracy even under arbitrary label assignments (e.g., 86.8\% on mini-ImageNet using WordNet IDs as labels), where zero-shot transfer fails. Additional results are reported in Appendix~\ref{app:image_analysis}.

\paragraph{Efficiency.}  
FAAST substantially reduces learning cost compared to backpropagation, saving approximately 95\% of GPU training time (see Appendix~\ref{app:image_settings}). It also outperforms memory-based methods in both accuracy and efficiency. Unlike retrieval approaches, which must store and access all key-value pairs at inference, FAAST compresses associative knowledge into a fixed-size fast weight matrix. As a result, both Linear Projection and FAAST incur $\mathcal{O}(d_x d_y)$ inference cost, whereas memory-based methods scale as $\mathcal{O}(N d_x + N d_y)$. FAAST achieves up to 90\% reduction in inference FLOPs and up to 95\% lower memory usage relative to memory-based methods, while maintaining superior accuracy.

\subsection{Text Classification}
Text classification tasks provide a discrete, semantic domain where we examine whether FAAST can serve as an efficient alternative to prompt-based adaptation.

\paragraph{Settings.}  
We evaluate FAAST on text classification (sentiment analysis) to assess its supervised learning capability, using \emph{GPT2-XL}~\citep{radford2019language} as the frozen backbone and following the standard integration procedure. We compare FAAST against \emph{zero-shot} inference and \emph{In-Context Learning (ICL)}~\citep{brown2020language}. Experiments are conducted on two benchmark datasets, \emph{SST-2}~\citep{socher2013recursive} and \emph{IMDB}~\citep{maas2011learning}. SST-2 contains 67{,}349 training examples and 1{,}821 test examples, while IMDB contains 25{,}000 training and 25{,}000 test examples. For each dataset, we randomly sample 5{,}000 training instances as the full support set and 5{,}000 test instances as the query set, except for SST-2, where all test examples are used for evaluation. We consider three adaptation regimes: 1-shot, 5-shot, and full-data adaptation.

\paragraph{Results.}  
Table~\ref{tab:gpt2_sentiment_results} summarizes sentiment classification performance on SST-2 and IMDB. Across both datasets and all adaptation regimes, FAAST substantially outperforms zero-shot inference and In-Context Learning. In low-shot settings, FAAST exhibits especially strong gains: on SST-2, FAAST improves accuracy from 59.6\% (ICL, 1-shot) to 78.5\%, and further to 80.8\% with 5-shot adaptation; similar trends are observed on IMDB, where FAAST achieves 86.7\% accuracy in the 1-shot setting, surpassing both zero-shot and ICL baselines by a large margin. Under full-data adaptation, FAAST consistently exceeds zero-shot GPT2-XL, reaching 87.5\% on SST-2 and 90.4\% on IMDB. These results demonstrate that FAAST enables effective supervised adaptation on top of frozen language models, achieving robust performance improvements. In this experiment, we confirm the basic capabilities on text, deferring more complex sequence modeling experiments and analyses to the next section.

\section{Experiments on Sequence Modeling Tasks}
We next evaluate FAAST on sequence modeling and conditional generation tasks, which impose stronger requirements on temporal credit assignment and long-range dependency handling. These experiments probe whether forward-only associative adaptation can operate at the sequence level, enabling test-time learning and cross-task generalization.

\begin{figure}[t]
    \centering
    \includegraphics[trim={0pt 0pt 0pt 0pt},clip,width=1.0\linewidth]{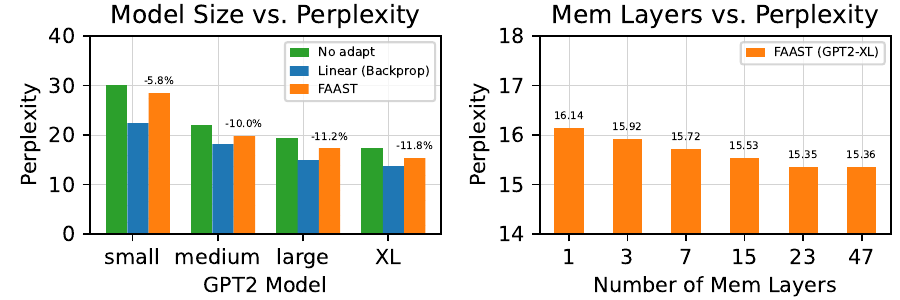}   
    \caption{GPT2 model size and mem layers vs. perplexity on WikiText-103.}
    \label{fig:gpt2_modelsize_barchart}
\end{figure}

\begin{table*}[t]
\centering\small
\caption{Machine translation on IWSLT2017, bold BLEU scores for statistical significance at $p<0.05$.}
\label{tab:qwen2_mt_results}
\begin{tabular}{lcccccccc}
\toprule
& \multicolumn{2}{c}{\bf En-De} 
& \multicolumn{2}{c}{\bf De-En}
& \multicolumn{2}{c}{\bf En-Fr}
& \multicolumn{2}{c}{\bf Fr-En} \\

\textbf{Method} 
& \textbf{$1$-shot} 
& \textbf{full}
& \textbf{$1$-shot} 
& \textbf{full}
& \textbf{$1$-shot} 
& \textbf{full}
& \textbf{$1$-shot} 
& \textbf{full}
\\

\midrule

Qwen2.5-3B-Instruct (zero shot)
& - & 23.22
& - & 32.92
& - & 30.56
& - & 39.24
\\

In-Context Learning
& 23.03 & -
& 32.33 & -
& 31.85 & - 
& 38.51 & -
\\

\textbf{FAAST (Ours)} 
& 23.35 & \bf 25.22 
& \bf 33.23 & \bf 36.40
& 31.12 & \bf 35.09 
& \bf 39.46 & \bf 42.47
\\

\midrule

Qwen2.5-7B-Instruct (zero shot)
& - & 25.53
& - & 34.69
& - & 34.82
& - & 41.40
\\

In-Context Learning
& 25.39 & -
& 35.70 & - 
& 35.45 & -
& 40.86 & -
\\

\textbf{FAAST (Ours)} 
& \bf 26.77 & \bf 27.75 
& 35.34 & \bf 37.10 
& 35.67 & \bf 37.08 
& \bf 42.08 & \bf 43.93
\\
\bottomrule
\end{tabular}
\end{table*}

\subsection{Sequence Modeling}
We use language modeling to study how FAAST supports test-time adaptation in autoregressive models.

\paragraph{Settings.}  
We evaluate FAAST across GPT-2 variants, ranging from small (117M) to XL (1.5B), following the standard integration procedure. Our method is compared with four primary baselines: \emph{zero-shot} without adaptation, \emph{Linear Projection} and \emph{LoRA} \cite{hu2022lora} with gradient-based adaptation, \emph{In-Context Learning}, and non-parametric \emph{kNN-LM} \cite{khandelwal2019generalization}. We evaluate language modeling performance on the \emph{WikiText-103} dataset \cite{merity2016pointer} following standard protocols. The entire training split is used for adaptation, while evaluation is performed on the held-out test split using \emph{perplexity (PPL)} as the metric. In certain experiments, the readout projection is additionally trained on the WikiText-103 training split to provide a controlled upper bound, reflecting performance when the readout has already seen the target-domain distribution.

\paragraph{Results.}
Table~\ref{tab:gpt2_ppl_results} shows that FAAST improves over the zero-shot GPT2-XL baseline with modest inference overhead. It achieves competitive perplexity against memory baselines with far lower memory usage, and when the readout projection is trained on in-domain data, perplexity drops to 13.23, matching or exceeding backprop-based adaptation. FAAST’s effectiveness grows with model size, giving up to 11.8\% relative perplexity reduction as illustrated by Figure~\ref{fig:gpt2_modelsize_barchart}, and is influenced by number of layers. Ablations show that the memory scorer and readout design provide the most efficient trade-off compared to more complex encoders or attention readouts (see Appendix~\ref{app:ablation_analysis} for details).

\subsection{Conditional Sequence-to-Sequence}
We use machine translation to evaluate whether FAAST generalizes to conditional sequence-to-sequence tasks, requiring structured input-output associations.

\paragraph{Settings.}  
We evaluate FAAST on machine translation using Qwen2.5-3B-Instruct and Qwen2.5-7B-Instruct \citep{qwen2.5}, following the standard integration procedure. Experiments are conducted on \emph{IWSLT2017} language pairs (En-De, De-En, En-Fr, Fr-En) \citep{cettolo2017overview} with 5,000 training samples as the support set and 5,000 test samples as the query set. We report \emph{BLEU} scores for both 1-shot and full-data adaptation scenarios.  

\paragraph{Results.}  
As shown in Table~\ref{tab:qwen2_mt_results}, FAAST consistently improves translation performance across all language pairs and adaptation settings. In particular, full-data adaptation with FAAST achieves substantial BLEU gains over zero-shot and In-Context Learning baselines. Specifically, it boosts De-En, En-Fr, and Fr-En translation with Qwen2.5-3B by more than 3 BLEU points and others by at least 2 BLEU points. These results demonstrate that forward-only adaptation extends effectively to structured text generation tasks, enabling efficient supervised adaptation.

\section{Discussion and Limitations}

FAAST demonstrates that many downstream adaptation problems can be efficiently addressed via associative mappings on top of frozen representations, bypassing both gradient-based optimization and context-length-dependent inference. By compiling labeled examples into task-specific fast weights, FAAST achieves rapid, single-pass adaptation with constant inference cost, offering a practical alternative to backpropagation and in-context learning. However, our approach relies on the quality of pretrained representations: if the frozen encoders do not capture task-relevant features, associative adaptation may fail. Furthermore, while FAAST excels at tasks with well-defined input–output correspondences, it may be less effective for problems requiring compositional reasoning, hierarchical dependencies, or long-range planning, where iterative or gradient-based refinement remains advantageous. Finally, current evaluations focus on classification and sequence modeling; extending FAAST to more complex structured prediction or multimodal tasks warrants future investigation.

\section{Conclusion}

We introduced FAAST, a forward-only associative adaptation paradigm that enables rapid task-specific learning without backpropagation or context-dependent inference. FAAST achieves efficient, single-pass adaptation with constant inference cost. Experiments across classification and sequence modeling tasks show that FAAST matches or exceeds the performance of backprop-trained and in-context learning baselines while substantially reducing computation and memory usage. Our results suggest that associative, gradient-free adaptation is a viable and practical alternative for deploying pretrained models in multi-task, online, or test-time adaptation scenarios. Future work includes extending FAAST to structured prediction, compositional reasoning, and multimodal tasks.

\section*{Impact Statement}
FAAST introduces a new paradigm for task adaptation that is both computationally efficient and memory-light, addressing practical bottlenecks in deploying large pretrained models across many downstream tasks. By eliminating gradient-based optimization and context-length-dependent inference, FAAST can substantially reduce energy consumption and hardware requirements, making adaptation accessible to resource-constrained settings. This approach also broadens the applicability of pretrained models to online, continual, and real-time learning scenarios where traditional training is infeasible. Beyond efficiency, FAAST provides a framework for understanding how associative memory principles can complement modern neural architectures, potentially inspiring future research on modular, interpretable, and scalable adaptation mechanisms.

% In the unusual situation where you want a paper to appear in the
% references without citing it in the main text, use \nocite
\nocite{langley00}

\bibliography{icml2026}
\bibliographystyle{icml2026}

%%%%%%%%%%%%%%%%%%%%%%%%%%%%%%%%%%%%%%%%%%%%%%%%%%%%%%%%%%%%%%%%%%%%%%%%%%%%%%%
%%%%%%%%%%%%%%%%%%%%%%%%%%%%%%%%%%%%%%%%%%%%%%%%%%%%%%%%%%%%%%%%%%%%%%%%%%%%%%%
% APPENDIX
%%%%%%%%%%%%%%%%%%%%%%%%%%%%%%%%%%%%%%%%%%%%%%%%%%%%%%%%%%%%%%%%%%%%%%%%%%%%%%%
%%%%%%%%%%%%%%%%%%%%%%%%%%%%%%%%%%%%%%%%%%%%%%%%%%%%%%%%%%%%%%%%%%%%%%%%%%%%%%%
\newpage
\appendix
\onecolumn

\section{Related Work}
\label{app:related_work}

While individual components of FAAST -- associative memory, fast weights, frozen representations, and pseudoinverse solutions -- have been studied in isolation, prior work does not simultaneously achieve forward-only learning, closed-form associative memory, non-parametric storage, and inference without memory access. FAAST occupies this previously unexplored combination, providing a new paradigm for modular and efficient downstream adaptation that is compatible with pretrained models, including large language models.

\subsection{Task Adaptation}

\paragraph{Parameter-efficient adaptation of pretrained models.}
Adapters \citep{houlsby2019parameter}, LoRA \citep{hu2022lora}, and prefix tuning \citep{li2021prefix} reduce the cost of downstream adaptation by introducing small trainable parameter sets. However, these methods still require gradient-based optimization. FAAST computes task-specific mappings analytically via forward-only associative memory, eliminating the need for any parameter training for downstream adaptation.

\paragraph{In-Context Learning and Test-Time Adaptation.}
In-context learning (ICL) in large language models enables task adaptation through conditioning on demonstrations provided at inference time \citep{brown2020language}. Subsequent work has studied ICL as implicit Bayesian inference or as retrieval over internal representations \citep{garg2022can,ahn2023transformers}. Unlike ICL, which requires storing and attending over the demonstration context at inference time, FAAST absorbs task-specific information into fast weights that can be reused across queries, offering a more memory- and computation-efficient alternative for test-time adaptation.

\subsection{Associative Learning}

\paragraph{Representation learning and associative learning.}
The separation between representation learning and task-specific association has appeared in multiple forms. Linear probes are often used to evaluate frozen representations \citep{alain2016understanding}, and world models separate feature learning from downstream prediction or control \citep{ha2018world}. Actor--critic methods also decouple value estimation from policy learning \citep{sutton1998reinforcement}. However, in most cases, gradients or task losses still influence the learned representations, or associative components are optimized using error-driven learning. In contrast, FAAST enforces a strict architectural separation: associative learning operates solely on fixed representations, without gradient flow or prediction-error signals.

\paragraph{Forward-only and biologically inspired learning rules.}
Several approaches explore alternatives to backpropagation, including local learning rules, feedback alignment, and forward-forward algorithms \citep{lillicrap2016random,hinton2022forward}. These methods often aim for biological plausibility or efficiency but usually require multiple passes or specialized training. FAAST is distinct: it targets downstream task adaptation on pretrained representations, achieving learning in a single forward pass with analytic guarantees. Related zeroth-order forward-only adaptation methods \citep{malladi2023fine} exist but rely on stochastic search rather than deterministic closed-form fast weights.

\paragraph{Fast weights and associative memories.}
Associative memory has a long history, from Hopfield networks \citep{hopfield1982neural,hopfield2007hopfield} and Hebbian learning \citep{hebb1949organization,kanter1987associative,personnaz1985information} to modern fast-weight models \citep{schmidhuber1992learning,ba2016using}. Traditional approaches rely on iterative updates or learned plasticity rules. FAAST differs by computing task-specific fast weights analytically from stored key-value pairs in a single forward pass, yielding deterministic, optimizer-free adaptation. Pseudoinverse-based associative memories \citep{pham2022generative} support high-fidelity retrieval but have not been combined with pretrained representations or inference-time compression.

\paragraph{Non-parametric memory and retrieval-augmented models.}
Memory-augmented neural networks, such as Neural Turing Machines \citep{graves2014neural}, Memory Networks \citep{weston2014memory}, Differentiable Neural Computers \citep{graves2016hybrid}, and modern Hopfield networks \citep{ramsauerhopfield}, enable rapid learning through attention-based reads and writes but require memory access during inference. Retrieval-based models, including kNN-LM \citep{khandelwal2020generalization} and RAG \citep{lewis2020retrieval}, also rely on querying stored key-value pairs at inference time. FAAST compresses all stored associations into a single fast-weight matrix, eliminating memory queries at inference while retaining the ability to adapt to new tasks.

\section{Theoretical Foundations}

This section provides theoretical justification for the design choices underlying FAAST.
We analyze fast weights as optimal solutions to linear regression, establish the necessity
of negative attention weights via pseudoinverse attention, justify the incremental update
rule for fast weights, and discuss alternative modular designs that preserve forward-only
computation.

\subsection{Fast Weights as Optimal Solutions to Regression}
\label{app:fastweights_regression}

Let $\{(\mathbf{k}_i, \mathbf{v}_i)\}_{i=1}^N$ denote a set of key--value pairs with
$\mathbf{k}_i \in \mathbb{R}^{d_x}$ and $\mathbf{v}_i \in \mathbb{R}^{d_y}$.
Let $K \in \mathbb{R}^{N \times d_x}$ and $V \in \mathbb{R}^{N \times d_y}$ be the
corresponding matrices.
Consider the linear predictor $W \in \mathbb{R}^{d_x \times d_y}$ defined by the
least-squares objective
\begin{equation}
\mathcal{L}(W) = \|K W - V\|_F^2 .
\end{equation}

\begin{theorem}[Optimality of Fast Weights]
\label{thm:fastweights_optimal}
The unique minimum-norm global minimizer of $\mathcal{L}(W)$ is
\begin{equation}
W^\star = K^\dagger V ,
\end{equation}
where $K^\dagger$ denotes the Moore--Penrose pseudoinverse of $K$.
Moreover, any gradient-based optimization method initialized at $W_0 = 0$ and using
a sufficiently small step size converges to $W^\star$.
\end{theorem}

This result shows that the fast weights computed by FAAST coincide exactly with the
solution obtained by gradient-based training of a linear predictor, but are obtained
analytically without gradients, stochasticity, or iterative updates. FAAST therefore
implements a fully compressed associative memory.

\subsection{Fast Weights as Pseudoinverse Attention}
\label{app:pseudoinverse_attn}

The closed-form solution $W^\star$ admits an interpretation as attention-based retrieval, analogous to Eq. \ref{eq:attention} in Section \ref{sec:preliminaries}.  
Fast weights can be viewed as a special form of attention \citep{vaswani2017attention} that retrieves key-value pairs through a least-squares criterion (see Appendix \ref{app:pseudoinverse_attn}). We term this mechanism \textbf{pseudoinverse attention}, which solves the exact retrieval problem
\begin{equation}
\min_{\mathbf{a}} \| K^\top \mathbf{a} - \mathbf{q} \|_2^2.
\end{equation}
The solution is given by $\mathbf{a}^\star = K^\dagger \mathbf{q}$, leading to the retrieved output
\begin{equation}
\mathbf{h} = (\mathbf{a}^\star)^\top V = \mathbf{q}^\top (K^\dagger V) = \mathbf{q}^\top W^\star.
\end{equation}
Unlike standard attention mechanisms, pseudoinverse attention permits attention weights $\mathbf{a}^\star$ to take negative values, reflecting a fundamentally different retrieval behavior.

\paragraph{Relation to Classic Attention Mechanisms.}
FAAST represents the fully compressed limit of attention-based memory by collapsing retrieval into a single fast-weight matrix, achieving maximal compression without inference-time memory access. Unlike $k$NN or softmax attention, which are restricted to discrete retrieval or convex combinations, FAAST utilizes pseudoinverse-based signed attention weights. This design choice enables both additive and subtractive interactions, allowing the module to represent a broader class of linear mappings. We justify the advantages of signed attention weights empirically in Appendix \ref{app:negative_attnweights}.

\paragraph{Softmax Attention as a Regularized Relaxation.}
The above formulation shows that pseudoinverse attention is the exact retrieval solution for the query $\mathbf{q}$. Now we show that softmax attention can be interpreted as an entropy-regularized relaxation of this optimal retrieval.

\begin{lemma}[Softmax Attention as an Entropy-Regularized Least-Squares Relaxation]
\label{lem:softmax_to_pinv}
Let $\mathbf{q} \in \mathbb{R}^{d_x}$ be a query and
$K \in \mathbb{R}^{N \times d_x}$ be keys.
Softmax attention computes
\[
\mathbf{a}_\tau
= \mathrm{softmax}\!\left(\frac{1}{\tau} K \mathbf{q}\right),
\qquad
\mathbf{h}_\tau = \mathbf{a}_\tau^\top V,
\]
where $\tau > 0$ is the temperature.
Then $\mathbf{a}_\tau$ is the solution to the entropy-regularized least-squares problem
\[
\min_{\mathbf{a}} \;\|K^\top \mathbf{a} - \mathbf{q}\|_2^2
\;-\; \tau \, H(\mathbf{a}),
\]
where
\[
H(\mathbf{a}) = -\sum_i a_i \log a_i,
\qquad
a \in \{ a \in \mathbb{R}^n \mid a_i \ge 0, \sum_i a_i = 1 \}.
\]
Moreover, as $\tau \to 0$ and provided $K$ has full column rank,
\[
\mathbf{a}_\tau \;\longrightarrow\; K^\dagger \mathbf{q},
\qquad
\mathbf{h}_\tau \;\longrightarrow\; (\mathbf{a}^\star)^\top V .
\]
\end{lemma}

\subsection{Incremental Update Rule for Fast Weights}
\label{app:incremental_update}

We analyze the incremental update of fast weights when new key--value pairs arrive
in batches, and show that this formulation naturally subsumes linear interpolation
of fast weights as a special case.

\paragraph{Batch-wise Formulation.}
Let an initial set of key--value pairs $(K_0, V_0)$ induce fast weights
\begin{equation}
W_0 = K_0^\dagger V_0 .
\end{equation}
Suppose a new batch of key--value pairs $(K_b, V_b)$ arrives.
The combined dataset is
\begin{equation}
K = \begin{bmatrix} K_0 \\ K_b \end{bmatrix},
\qquad
V = \begin{bmatrix} V_0 \\ V_b \end{bmatrix}.
\end{equation}

The fast weights corresponding to the combined memory are given by the least-squares
solution
\begin{equation}
W^\star = \arg\min_W \; \|K W - V\|_F^2
= (K^\top K)^{-1} K^\top V ,
\end{equation}
assuming $K^\top K$ is invertible.

\paragraph{Incremental Update via Sufficient Statistics.}
Define the sufficient statistics
\begin{equation}
S = K^\top K,
\qquad
T = K^\top V .
\end{equation}
When a new batch $(K_b, V_b)$ is added, these statistics update additively:
\begin{equation}
S \leftarrow S + K_b^\top K_b,
\qquad
T \leftarrow T + K_b^\top V_b .
\end{equation}

Using the Sherman--Morrison--Woodbury identity, the inverse $S^{-1}$ can be updated
efficiently without recomputing from scratch, yielding an exact update of
\begin{equation}
W^\star = S^{-1} T .
\end{equation}
This establishes FAAST as an exact, forward-only method for batch-wise online
associative learning.

\paragraph{Linear Interpolation as an Approximation.}
Consider two disjoint batches $(K_1, V_1)$ and $(K_2, V_2)$ with corresponding
sufficient statistics $(S_1, T_1)$ and $(S_2, T_2)$.
Let
\begin{equation}
S = \lambda S_1 + (1-\lambda) S_2,
\qquad
T = \lambda T_1 + (1-\lambda) T_2,
\end{equation}
for some $\lambda \in [0,1]$.
Then the resulting fast weights satisfy
\begin{equation}
W = S^{-1} T
= \lambda W_1 + (1-\lambda) W_2,
\end{equation}
when $S_1$ and $S_2$ are mutually orthogonal or proportional to the identity. Here, $W_i = S_i^{-1} T_i$.

Thus, when global attention weights or task contributions combine linearly across batches, the induced fast weights interpolate linearly as well. This property explains why FAAST supports smooth task interpolation and mixture-of-tasks behavior without catastrophic interference.

\paragraph{Discussion.}
The batch-wise incremental formulation unifies online updates and task interpolation within a single least-squares framework. FAAST therefore supports continual, multi-task, and test-time adaptation using a single deterministic update rule, without gradients or iterative optimization.

\section{Experimental Setup}

All experiments are conducted on NVIDIA H100 GPUs with 80 GB memory. Image classification models are trained and evaluated on a single GPU, while LLMs are trained using 8 GPUs and evaluated on a single GPU.

\subsection{Image Classification Settings}
\label{app:image_settings}

\paragraph{Base Model.}
We choose pretrained CLIP ResNet-50 \citep{radford2021learning} as the backbone model, using frozen image and text encoders. Image embeddings serve as keys, and text embeddings of class prompts ``{\it A photo of a \{label\}.}'' serve as values. All adaptation is performed on these fixed representations.

\paragraph{Baselines.}
All methods operate on identical frozen features to isolate the effect of associative memory. \emph{CLIP zero-Shot} makes prediction using cosine similarity between image and text embeddings \citep{goh2021multimodal}. \emph{Linear projection (backprop)} trains a linear classifier \citep{kolesnikov2019revisiting} using stochastic gradient descent with momentum.  \emph{Full finetuning (backprop)} trains a linear classifier together with the image encoder. \emph{k-NN memory}  does nearest-neighbor retrieval \citep{wu2018unsupervised} with $k=\min(n,10)$. \emph{Softmax memory} does attention-based retrieval \citep{vaswani2017attention}. For k-NN, softmax memory, and FAAST, predictions are linearly interpolated with CLIP zero-shot predictions using the same prior count $N_0$. We set $N_0$ to 40 times the number of classes, yielding $N_0 = 400$ for CIFAR-10 and $N_0 = 800$ for mini-ImageNet.
All other hyperparameters are fixed across datasets, including $\alpha=0.8$ for singular value filtering.

\paragraph{Datasets and Evaluation.}
We evaluate on CIFAR-10 \citep{krizhevsky2009learning} and mini-ImageNet \citep{vinyals2016matching} datasets. \emph{CIFAR-10} contains 10 classes with 50{,}000 training and 10{,}000 test images; we use the training split as support set and the test split as query set.  
\emph{mini-ImageNet} contains 100 classes. We use the 20-class test split only, randomly dividing each class into equal size, obtaining 6{,}000 samples for support set and another 6{,}000 for query set.

We evaluate the methods under few-shot and full-data settings.
\emph{Few-shot} evaluation follows a $k$-way $n$-shot episodic protocol \citep{vinyals2016matching}. In each episode, $n$ labeled samples per class are drawn from the support set to construct the classifier, and 20 query samples per class are used for evaluation. Results are averaged over 600 episodes with 95\% confidence intervals.  
For \emph{full-data} evaluation, all support samples are used for learning or memory construction, and all query samples are used for testing.

\paragraph{Backprop Model Training.}
Table \ref{tab:image_classification_settings} summarizes the training configurations used for backpropagation-based baselines in image classification. Both linear projection and full finetuning are built on a CLIP ResNet-50 backbone. Linear projection trains only a linear classifier on top of frozen image features, while full finetuning additionally updates the image encoder. We use SGD with momentum, a shared learning rate schedule, and identical batch size and regularization to ensure a controlled comparison across methods.

\begin{table*}[h]
\begin{minipage}{0.4\linewidth}
% \begin{table}[h]
\caption{Image Classification Training Settings.}
\label{tab:image_classification_settings}
\centering\scriptsize
\begin{tabular}{ccc}
\toprule
\bf Argument & \bf Linear Projection & \bf Full Finetuning \\
\midrule
num epochs & 20 & 30 \\
learning rate & 1.0e-03 & 1.0e-05 \\
base model & \multicolumn{2}{c}{CLIP resnet-50}  \\
dropout & \multicolumn{2}{c}{0.3} \\
optim & \multicolumn{2}{c}{SGD(momentum=0.9)} \\
lr scheduler type & \multicolumn{2}{c}{StepLR(step size=10, gamma=0.1)} \\
warmup steps & \multicolumn{2}{c}{500} \\
batch size & \multicolumn{2}{c}{16} \\
GPUs & \multicolumn{2}{c}{1} \\
torch dtype & \multicolumn{2}{c}{float32} \\
\bottomrule
\end{tabular}
% \end{table}

\end{minipage}\hfill
\begin{minipage}{0.58\linewidth}
% \begin{table*}[h]
\caption{Large Language Model Training Settings.}
\label{tab:llm_settings}
\centering\scriptsize
\begin{tabular}{cccc}
\toprule
\bf Argument & \bf Linear & \bf LoRa & \bf Readout \\
\midrule
base model & \multicolumn{3}{c}{ GPT2, GPT2-Medium, GPT2-Large, GPT2-XL} \\
& & & Qwen2.5-3B/7B-Instruct \\
optim & \multicolumn{3}{c}{AdamW} \\
learning rate & \multicolumn{3}{c}{2.0e-05} \\
lr scheduler type & \multicolumn{3}{c}{cosine} \\
max length & \multicolumn{3}{c}{1024} \\
batch size & \multicolumn{3}{c}{8} \\
GPUs & \multicolumn{3}{c}{8} \\
gradient checkpointing & \multicolumn{3}{c}{true} \\
gradient accumulation steps & \multicolumn{3}{c}{1} \\
num train epochs & \multicolumn{3}{c}{1} \\
warmup steps & \multicolumn{3}{c}{500} \\
packing & \multicolumn{3}{c}{true} \\
packing strategy & \multicolumn{3}{c}{wrapped} \\
attn implementation & \multicolumn{3}{c}{flash attention 2} \\
torch dtype & \multicolumn{3}{c}{bfloat16} \\

\bottomrule
\end{tabular}
% \end{table*}

\end{minipage}
\end{table*}

\subsection{Language Modeling Settings}
\label{app:llm_settings}

\paragraph{Base Models.}  
We evaluate FAAST on a range of pretrained language models, including GPT-2 variants from small (117M) to XL (1.5B) \citep{radford2019language}, and Qwen2.5 instruct models at 3B and 7B scales \citep{qwen2.5}. All pretrained parameters remain frozen throughout adaptation, which ensures that improvements can be attributed solely to FAAST’s forward-only adaptation mechanism.

\paragraph{Baselines.}  
We compare FAAST to four categories of baselines. \emph{zero shot} serves as a non-adaptive lower bound, establishing the baseline performance without any sequence-level adaptation. \emph{Linear projection} and \emph{LoRa} \citep{hu2022lora} trained via backpropagation on frozen base model using cross-entropy loss, represents standard gradient-based adaptation. \emph{In-Context Learning (ICL)} \citep{brown2020language} prepends demonstration examples to the input sequence at inference time, enabling forward-only adaptation without updating any model parameters. \emph{kNN-LM} \citep{khandelwal2019generalization} is a non-parametric memory baseline that stores all pairs of hidden states and output tokens from the training split, providing a direct comparison to FAAST’s memory-based mechanism. All methods use the same adaptation examples, and for ICL, the number of demonstrations is matched to the number of examples stored in FAAST’s memory.

\paragraph{Readout Pretraining.}  
FAAST introduces two key components for sequence-level adaptation: a \emph{token scorer}, which identifies informative tokens to store in memory, and a \emph{memory readout projection}, which interprets the stored memory items during prediction. Unless specified otherwise, both components are pretrained jointly on 1\% of OpenWebText2 \footnote{https://huggingface.co/datasets/segyges/OpenWebText2}. During training, to avoid the dominance of historical fast weights computed using outdated readout and weighting, we apply a discount to incremental update $N_t$ before each update, with an empirical value of 0.9.

\paragraph{Backprop Model Training.}

Table~\ref{tab:llm_settings} summarizes the training configurations for backpropagation-based baselines in language modeling. Both linear projection and LoRA fine-tuning are applied on top of a frozen base model. Linear projection trains a residual-connected linear probe inserted between Transformer layers, while LoRA fine-tunes attention and FFN weights with a matched number of trainable parameters. All baselines are optimized using AdamW with a cosine learning-rate scheduler. For smaller models (GPT2, GPT2-Medium, and GPT2-Large), a projection is inserted after every layer; for larger models (GPT2-XL, Qwen2.5-3B-Instruct, and Qwen2.5-7B-Instruct), projections are inserted every two layers. At the same insertion locations, we add a residual-connected FAAST module for our method.

\section{Experiments}

\subsection{Image Classification Results and Analysis}

\subsubsection{Training and Learning Cost Comparison}

\begin{table}[h]
\caption{Image Classification Training/Learning Costs.}
\label{tab:image_classification_costs}
\centering\small
\begin{tabular}{lcc}
\toprule
\bf Method & \bf Train/Learn Cost $\downarrow$ \\

\midrule
\multicolumn{3}{l}{\it On CIFAR10 with 50,000 training and 10,000 test samples:} \\
Full finetuning (backprop) & 1,512 GPU seconds \\
Linear projection (backprop) & 568 GPU seconds \\
\bf FAAST (ours) & \bf 38 GPU seconds (-93.3\%) \\

\midrule
\multicolumn{3}{l}{\it On mini-ImageNet with 6,000 training and 6,000 test samples:} \\
Full finetuning (backprop) & 212 GPU seconds \\
Linear projection (backprop) & 91 GPU seconds \\
\bf FAAST (ours) & \bf 3 GPU seconds (-96.7\%) \\

\bottomrule
\end{tabular}
\end{table}

Table \ref{tab:image_classification_costs} reports the corresponding training or learning cost measured in GPU seconds. Backpropagation-based methods incur substantial computational overhead due to iterative optimization over multiple epochs. In contrast, FAAST performs a single-pass, closed-form associative update without gradient computation, optimizer state, or repeated epochs. As a result, FAAST reduces learning time by over 93\% on CIFAR-10 and over 96\% on mini-ImageNet, while operating on the same frozen pretrained representations.

\subsubsection{Necessity of Negative Attention Weights}
\label{app:negative_attnweights}

We justify the use of negative attention weights as follows and support it by empirical experiments.

\paragraph{Linear Attention with Negative Attention Weights.}
Standard linear attention compute retrieval weights based on kernel similarity
scores between the query $\mathbf{q}$ and keys $\mathbf{k}_i$. Without any adjustment,
it often produce low-contrast attention, failing to sharply differentiate
relevant from irrelevant memory items. In memory retrieval experiments, such as image
classification, standard linear attention give poor performance.

Linear attention allows partial precomputation of memory statistics.
However, due to its linearity and non-negativity constraints on kernel function $\varphi(\cdot)$,
linear attention has limited ability to suppress irrelevant keys.
We improve it by centralizing the similarity scores computed with kernelized inner products:
\begin{equation}
\label{eq:linear_attn}
a_i
= \frac{s_i - \bar{s}}{\sum_j s_j},
\qquad
s_i = \varphi(\mathbf{q})^\top \varphi(\mathbf{k}_i).
\end{equation}
This centralization enhances the
contrast of attention weights, allowing relevant keys to be emphasized more strongly
relative to irrelevant ones. Empirically, this modification significantly improves
memory retrieval performance.

After centralization, linear attention weights in Eq. \ref{eq:linear_attn}
can become negative for irrelevant keys. These negative weights effectively suppress
irrelevant memory items, performing a type of null-space cancellation. In other words,
negative attention weights allow linear attention to selectively retrieve relevant content, which is otherwise impossible with strictly non-negative weights.

\begin{table*}[h]
\centering\small
\caption{Necessity of Negative Attention Weights.}
\label{tab:negative_attnweights_results}
\begin{tabular}{lc@{\hspace{40pt}}c@{\hspace{40pt}}c}
\toprule
\multirow{2}{*}{\bf Method} 
& \multicolumn{3}{c}{\bf {\it mini}ImageNet (20-way with WordNet ID as Label)}  \\
 
& 1-shot 
& 5-shot 
& Full-Set \\

\midrule
Linear attention memory
& 5.0$\pm$0.0 
& 5.0$\pm$0.0 
& 5.0$\pm$0.0 \\

Linear attention w/ negative weights
& 32.3$\pm$4.3 
& 37.6$\pm$2.9 
& 39.6$\pm$1.2 \\

\midrule
\bf FAAST (Ours)
& \bf 40.4$\pm$4.1 
& \bf 66.7$\pm$2.7 
& \bf 86.8$\pm$0.9 \\
\bottomrule
\end{tabular}
\end{table*}

\paragraph{Empirical Results.}  
Table~\ref{tab:negative_attnweights_results} presents the impact of allowing negative attention weights in linear attention on \emph{mini}ImageNet classification. Standard linear attention without negative weights fails to distinguish relevant memory items, resulting in near-random accuracy across all settings (1-shot, 5-shot, and full-set). Introducing negative weights via centralization substantially improves performance, confirming that suppressing irrelevant keys is crucial for effective memory retrieval. Our FAAST model further amplifies this effect, achieving the highest accuracy in all scenarios by combining negative-weighted linear attention with fast adaptation, demonstrating that selective retrieval of relevant memory items is essential for few-shot and full-set classification.

\subsubsection{Analysis of Generalization}
\label{app:image_analysis}

\begin{figure}[h]
    \centering
    \includegraphics[trim={0pt 0pt 0pt 0pt},clip,width=0.5\linewidth]{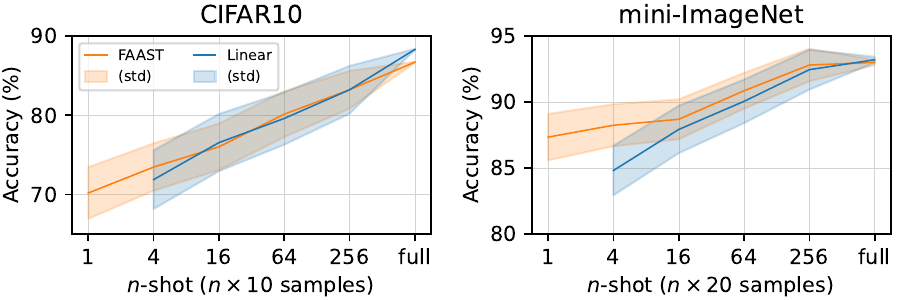}   
    \caption{FAAST vs. Linear (backprop). The std represents the variance of accuracy across episodes.}
    \label{fig:image_nshot_linechart}
\end{figure}

\begin{figure}[t]
    \centering
    \includegraphics[trim={0pt 0pt 0pt 0pt},clip,width=0.5\linewidth]{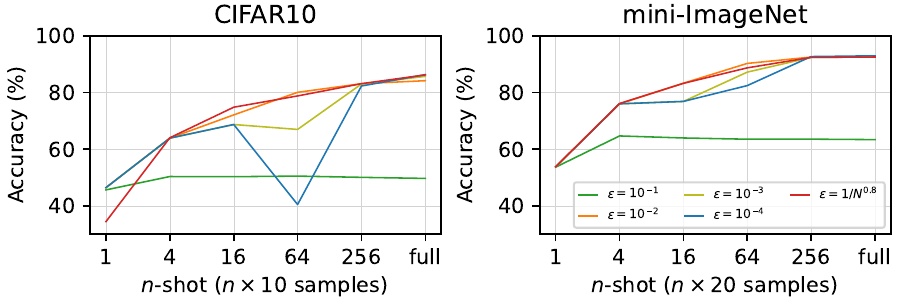}   
    \caption{Filter noisy components under a threshold $\epsilon$. Experiments are conducted with $N_0=0$ to avoid the influence of prior.}
    \label{fig:image_rtol_linechart}
\end{figure}

\paragraph{Generalization in Few-Shot Settings.}  
FAAST demonstrates a clear advantage in few-shot scenarios, where backpropagation-based training often suffers from severe overfitting due to limited data. As illustrated in Figure~\ref{fig:image_nshot_linechart}, FAAST matches or outperforms linear projections trained with backpropagation under 256-shot settings, while exhibiting lower variance in accuracy across runs. This highlights FAAST’s robustness and its ability to generalize effectively from small numbers of examples.

\paragraph{Generalization to Arbitrarily Defined Labels.}
CLIP relies on a pretrained semantic alignment between image and text embeddings. When this alignment is broken by assigning arbitrary class names, zero-shot performance degrades to near chance. On mini-ImageNet, using WordNet IDs (e.g., {\it n02119789}) as class names yields 6.4\% accuracy on the full data, close to the 5\% random baseline for 20-way classification.
In contrast, FAAST does not rely on prior semantic alignment. Without prior ($N_0=0$), it achieves 85.1\% accuracy in this setting, demonstrating strong generalization even when class labels are arbitrarily defined.

\paragraph{Overfitting-Underfitting Trade-off.}
FAAST computes fast weights via the pseudoinverse \citep{penrose1955generalized}, which relies on an SVD of the key matrix. Large singular values correspond to shared, generalizable components, whereas small singular values capture sample-specific variations. By filtering singular values using a relative tolerance threshold, we explicitly control the balance between memorization and generalization.
As shown in Figure~\ref{fig:image_rtol_linechart}, accuracy exhibits a function of the threshold given a fixed number of samples: overly aggressive filtering leads to underfitting, while insufficient filtering leads to overfitting. In practice, a relative tolerance of $\epsilon=1/N^{0.8}$ provides a robust balance and is used throughout the experiments.

\subsection{Language Modeling Results and Analysis}
\label{app:ablation_analysis}

\subsubsection{Training and Learning Cost Comparison}

\begin{table*}[h]
\centering\small
\caption{Large Language Model Train and Learn Cost Comparison.}
\label{tab:gpt2_lm_traincost}
\begin{tabular}{lccc}
\toprule
\textbf{Method} 
& \textbf{Adaptation} 
& \bf Train/Learn Cost
& \bf Inference Cost \\

\midrule
Linear Projection (upper bound)
& Backprop
& 3 GPU hours
& 23 GPU seconds \\

LoRa w/ same number of params
& Backprob
& 3 GPU hours
& 23 GPU seconds \\

kNN-LM
& Memory
& 16 GPU hours
& 660 GPU seconds \\

\bf FAAST (ours)
& Fast weights 
& \bf 0.2 GPU hrs (-93.3\% vs Backprop)
& \bf 23 GPU secs (-96.5\% vs Memory) \\

\bottomrule
\end{tabular}
\end{table*}

Table~\ref{tab:gpt2_lm_traincost} compares computational costs across adaptation methods. Backpropagation-based methods (e.g., LoRA) require moderate training ($\approx$3 GPU hours) and fast inference (23 GPU seconds), while memory-based approaches like kNN-LM are significantly more expensive in both phases. In contrast, FAAST reduces training to 0.2 GPU hours, saving 93.3\% over backpropagation, while matching the fastest inference speed. This represents a 96.5\% reduction in inference latency compared to memory-based models, making FAAST ideal for resource-constrained or latency-sensitive applications.

\subsubsection{Analysis of Influencing Factors}

\paragraph{Impact of Model Size and Number of Layers.}
We also examine the effect of model size and the number of memory layers. As Figure \ref{fig:gpt2_modelsize_barchart} shows, FAAST consistently improves the base models across all GPT-2 sizes, with relative perplexity reductions increasing from 5.8\% to 11.8\% as model size grows. Increasing the number of memory layers generally reduces perplexity, although gains plateau beyond a moderate depth, indicating diminishing returns. Accordingly, for smaller language models such as GPT2, GPT2-Medium, and GPT2-Large, we set the number of memory layers equal to the total number of Transformer layers. For larger models, we set the number of memory layers to half of the total Transformer layers.

\begin{figure}[h]
    \centering
    \includegraphics[trim={0pt 0pt 0pt 0pt},clip,width=0.5\linewidth]{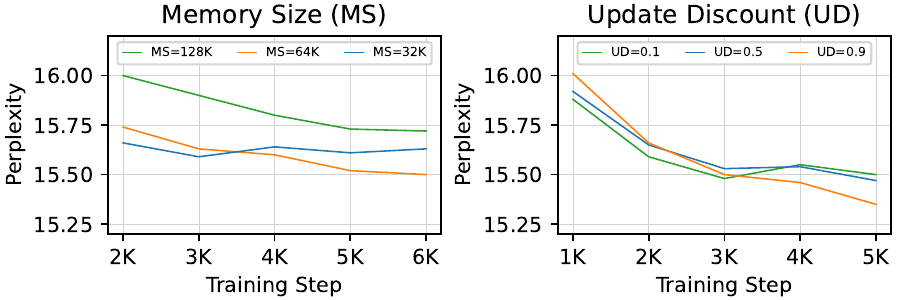}   
    \caption{Training dynamics for different memory size and update discount settings}
    \label{fig:gpt2_hyperparam_linechart}
\end{figure}

\paragraph{Impact of Hyperparameters.}
As Figure~\ref{fig:gpt2_hyperparam_linechart} illustrates, the maximum memory size (ranging from 32K to 128K tokens) governs both memory consumption and the frequency of fast weight updates. Fast weights are updated -- and the oldest batch of memory items removed -- only when the memory reaches its maximum capacity. The update discount plays a crucial role in preventing older fast weights from dominating the model. During training, as memory grows, the relative contribution of newly added batches diminishes. To mitigate the influence of outdated fast weights computed with stale readout projections and token scorers, it is necessary to progressively decay the effect of historical fast weights. Empirically, we use a maximum memory size of 64K and a update discount of 0.9 in the main experiments.

\subsubsection{Architectural Variants and Extensions}
\label{app:alternative_modules}

We further consider alternative module designs for target encoder and readout projection as follows.

\paragraph{Right-to-Left Encoder.}  
Let $x_i$ denote the left context and $y_i$ the right context. We introduce a Transformer layer with a right-to-left causal attention mask, stacked on top of the middle layers of the base model, to encode $y_i$. This layer is trained jointly with the readout projection and injects future context into the associative memory. Empirically, this design yields small but consistent improvements, reducing GPT2-XL's perplexity on WikiText-103 from 15.35 to 15.15. Due to the additional parameter overhead, we do not employ this in our main experiments.

\paragraph{Attention-Based Readout.}  
Instead of a linear readout that accesses only the current memory output, we use an attention-based readout layer that attends to all previous memory outputs. This enables aggregation of longer-range memory signals before prediction and is trained jointly with other components. This approach provides modest improvements, lowering perplexity by roughly 0.08, suggesting that local memory interpretation is often sufficient. Combining it with a right-to-left Transformer encoder for future-context encoding further reduces perplexity by about 0.20, but the parameter overhead prevents its use in the main experiments.

\end{document}